\documentclass[11pt]{article}

\usepackage[final]{acl}
\usepackage{hyperref}
\usepackage{url}
\usepackage{amsmath}
\usepackage[utf8]{inputenc} 
\usepackage[T1]{fontenc}    
\usepackage{hyperref}       
\usepackage{url}            
\usepackage{booktabs}       
\usepackage{amsfonts}       
\usepackage{nicefrac}       
\usepackage{microtype}      
\usepackage{xcolor}         
\usepackage{tikz} 
\usepackage{caption}
\usepackage{algorithm}
\usepackage{algorithmic}
\usepackage{graphicx}
\usepackage{amsmath}
\usepackage{caption}
\usepackage{graphicx}
\usepackage{multirow}
\usepackage{subcaption}
\usepackage[table]{xcolor}
\usepackage{times}
\usepackage{latexsym}

\usepackage[T1]{fontenc}

\usepackage[utf8]{inputenc}

\usepackage{microtype}

\usepackage{inconsolata}

\usepackage{graphicx}

\title{BiMol-Diff: A Unified Diffusion Framework for Molecular Generation and Captioning}

\author{
\textbf{Aditya Hemant Shahane}$^{1}$\thanks{Equal contribution} \quad
\textbf{Anuj Kumar Sirohi}$^{1}$\footnotemark[1] \quad
\textbf{Devansh Arora}$^{2}$ \\
\textbf{Nitin Kumar}$^{2}$ \quad
\textbf{Prathosh A P}$^{3,4}$ \quad
\textbf{Sandeep Kumar}$^{1}$ \\
$^{1}$Indian Institute of Technology Delhi, 
$^{2}$Indian Institute of Technology Ropar \\
$^{3}$Indian Institute of Science Bengaluru, 
$^{4}$Latentforce.ai
}

\begin{document}
\maketitle
\begin{abstract}

Bridging molecular structures and natural language is essential for controllable design. Autoregressive models struggle with long-range dependencies, while standard diffusion processes apply uniform corruption across positions, which can distort structurally informative tokens. We present \texttt{BiMol-Diff}, a unified diffusion framework for the paired tasks of text-conditioned molecule generation and molecule captioning. Our key component is a \textit{Token-aware noise schedule} that assigns position-dependent corruption based on token recovery difficulty, preserving harder-to-recover substructures during the forward process. On ChEBI-20 and M3-20M, \texttt{BiMol-Diff} improves molecule reconstruction with a 15.4\% relative gain in Exact Match and achieves strong captioning results, attaining the best BLEU and BERTScore among compared baselines. These results indicate token-aware noising improves fidelity in molecular structure--language modeling.
Code link \href{https://github.com/adityashahane10/BiMol-Diff-Unified-Diffusion-Framework-for-Molecular-Generation-and-Captioning.git}{GitHub}.

\end{abstract}

\section{Introduction}\label{Introduction}
Designing molecules requires balancing two distinct modalities: the molecular \emph{structure} of atoms and bonds, and the linguistic descriptions of molecular properties (e.g., ``a kinase inhibitor with good solubility''). Bridging this gap through cross-modal translation enables chemically grounded workflows where designs can be iteratively refined via natural language~\cite{zheng2023structureinformedlanguagemodelsprotein,fatemi2024talk}. This motivates two complementary directions: (i) \textbf{molecule generation} (\emph{text}$\rightarrow$\emph{molecule}) to materialize descriptions into structures, and (ii) \textbf{molecule captioning} (\emph{molecule}$\rightarrow$\emph{text}) to explain molecular structure and properties via natural-language descriptions~\cite{edwards-etal-2022-translation,zeng2023interactivemoleculardiscoverynatural}. However, despite their symmetry, these tasks are typically developed in isolation with separate models and objectives, limiting consistency and reuse in iterative design loops.

\noindent A common bridge between molecular structure and sequence models is the Simplified Molecular-Input Line-Entry System (SMILES)~\cite{doi:10.1021/ci00057a005,doi:10.1021/ci00062a008}, which linearizes a molecular graph into a string. This linearization has enabled a large family of sequence-based molecular methods, including (1) conditional SMILES generation from text prompts~\cite{doi:10.1021/acs.jcim.1c00600,D1RA03086H,irwin2022chemformer}, and (2) SMILES or structure-conditioned captioning that maps molecules to natural-language descriptions~\cite{edwards-etal-2022-translation,zeng2023interactivemoleculardiscoverynatural}. Many of these systems rely on \emph{autoregressive} (AR) pretrained language models (PLMs) such as GPT~\cite{10.1007/s11023-020-09548-1}, T5~\cite{JMLR:v21:20-074}, and BART~\cite{lewis-etal-2020-bart}, which generate tokens sequentially by conditioning each next token on the prefix~\cite{doi:10.1021/acs.jcim.1c00600,D1RA03086H,irwin2022chemformer}. In this work, while we retain SMILES as the standard input/output interface, we explicitly project these molecules into a Knowledge Graph (KG) view represented as a serialized sequence of atom-bond-atom triplets to achieve better generalization and reduce dependency on specific SMILES syntax rules (e.g., ring number). 

\noindent While effective, autoregressive models are not always well suited to SMILES-centric molecule--language modeling. 
First, decoding is inherently left-to-right, so early mistakes can propagate and are difficult to correct~\cite{nie2025largelanguagediffusionmodels,arriola2025blockdiffusioninterpolatingautoregressive}. 
Second, SMILES encodes long-range syntactic and chemical dependencies (e.g., ring closures, branching, and valid substructures) that require coordinated decisions across distant positions; locally plausible token choices can therefore cascade into globally invalid molecules. 
Third, many objectives are inherently structure-level such as scaffold preservation that are hard to enforce under strict prefix conditioning. 
These limitations motivate non-autoregressive generation mechanisms that can revise all positions jointly, raising a broader question: \emph{can we build models that generate and explain molecules while explicitly preserving global validity and structural controllability?}

\noindent Diffusion language models offer a promising non-autoregressive alternative by iteratively denoising a corrupted representation, enabling holistic updates over all positions at each step and better coordination of long-range dependencies~\cite{li2022diffusionlmimprovescontrollabletext,nie2025largelanguagediffusionmodels}. However, standard diffusion applies noise uniformly across tokens, which can corrupt chemically critical tokens (e.g., ring indices and branching markers) too early in the process and hinder recoverability.

\noindent To address this, we propose \texttt{BiMol-Diff}, a unified diffusion framework that addresses the paired tasks of \emph{text}$\rightarrow$\emph{SMILES} generation and \emph{SMILES}$\rightarrow$\emph{text} captioning under a common denoising formulation. Unlike standard diffusion, \texttt{BiMol-Diff} employs a token-aware noising strategy that assigns token-dependent noise levels using the per-token training loss as a proxy for recovery difficulty. Intuitively, tokens that are consistently harder to denoise receive a more conservative corruption schedule, improving recoverability. We apply this mechanism consistently across both directions: for \emph{SMILES}$\rightarrow$\emph{text}, it helps preserve semantic tokens during denoising, and for \emph{text}$\rightarrow$\emph{SMILES}, it helps preserve chemically salient SMILES tokens during generation.

\noindent \textbf{Contributions.} We make three primary contributions: (1) a token-aware noising strategy which preserves critical structural semantics during the diffusion process; (2) a unified framework that solves the bidirectional tasks of molecule generation and captioning, utilizing the same token-aware mechanism for both inverse problems; and (3) strong empirical performance across a wide range of metrics on both molecule generation and molecule captioning benchmarks.
\section{Background and Preliminaries} \label{BRW}

\subsection{Related Work}
\noindent \textbf{Diffusion for Molecular Graphs (Unconditional / Property-based):}
Diffusion models are widely used for molecular graph generation, spanning unconditional sampling and property-conditioned design. Recent variants improve expressivity via transformer-style graph denoisers \citep{liu2024graphdit}, explore hierarchical/latent formulations for scalability and structure preservation \citep{bian2024hglatentdiff}, and study control mechanisms for property prediction and guidance \citep{zhang2025mgdiff}.

\noindent \textbf{Text-guided Molecule Generation:}
Text-to-molecule generation is typically formulated as learning $p(\tilde{\mathcal{G}}\mid \mathbf{S})$ (or $p(\text{SMILES}\mid \mathbf{S})$), where $\mathbf{S}$ describes desired structure or function. Recent work adapts PLMs to chemical strings, enabling text$\leftrightarrow$molecule translation with encoder--decoder pretraining \citep{edwards-etal-2022-translation} and improved multi-task / chemistry-aware extensions \citep{kim2025camt5}. In parallel, diffusion-based text conditioning has emerged, including diffusion-LM style text-guided generation in discrete/embedded sequence spaces \citep{gong2024tgmdlm} and graph/latent diffusion variants that preserve molecular structure \citep{chang2025ldmoltexttomoleculediffusionmodel}.

\noindent \textbf{Molecule Captioning / Molecule-to-Text:}
Generating faithful descriptions from molecules (G2S) has been explored via translation-style baselines \citep{edwards-etal-2022-translation} and molecule captioning models that incorporate structural encoders to improve grounding \citep{liu2024molcamoleculargraphlanguagemodeling}. More recent multimodal approaches further strengthen molecule--language alignment for captioning and related tasks.

\noindent \textbf{Unifying / Bidirectional Molecule--Text Modeling:}
While text-guided generation and captioning have progressed independently, \emph{fully bidirectional} modeling that supports both $p(\mathbf{S}\mid \tilde{\mathcal{G}})$ and $p(\tilde{\mathcal{G}}\mid \mathbf{S})$ within a single generation framework remains relatively limited. Translation-based PLM systems offer practical bidirectionality \citep{edwards-etal-2022-translation}, and recent foundation-model efforts move toward broader cross-modal generalization \citep{liu2024molcamoleculargraphlanguagemodeling}, but diffusion-based approaches are typically developed for one direction (most often S2G) and do not explicitly address token-level corruption heterogeneity. \texttt{BiMol-Diff} fills this gap by casting \emph{both} directions under a unified conditional diffusion view and introducing a molecular graph-aware, token-dependent noising schedule to preserve chemically salient information during generation (see Appendix~\ref{related work}).

\subsection{Preliminaries}
\begin{figure*}[t!]
    \centering
    \includegraphics[width=0.8\linewidth]{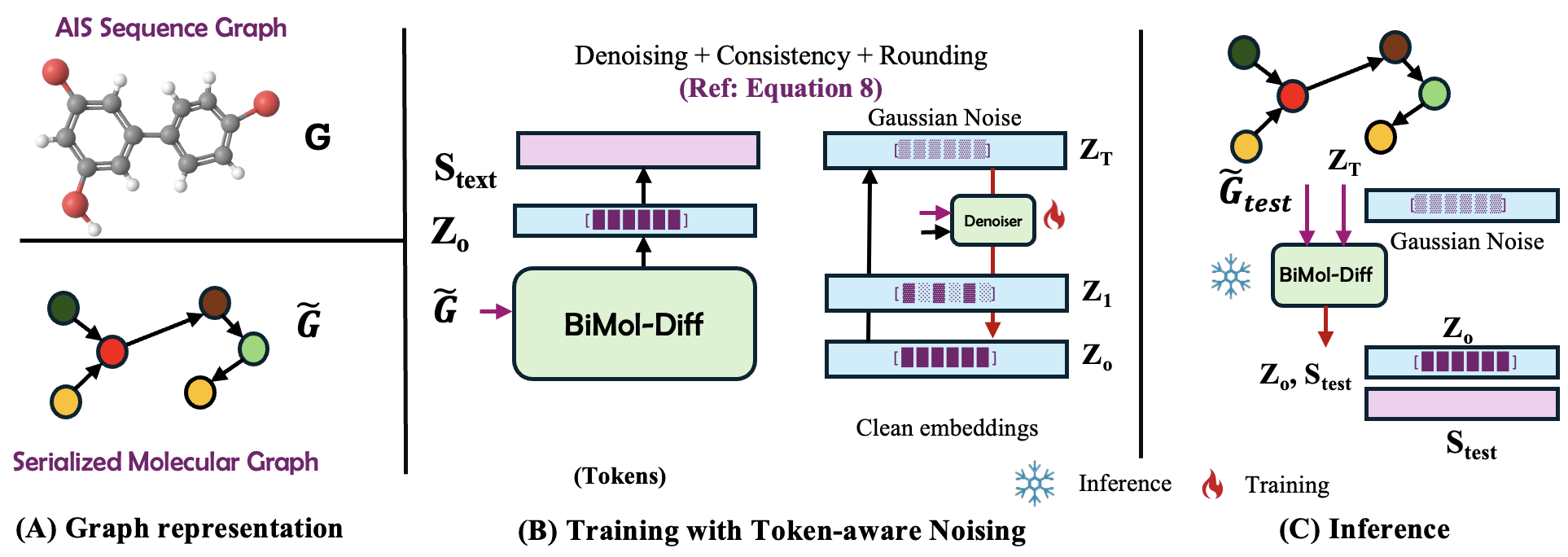}
    \caption{\textbf{The \texttt{BiMol-Diff} Framework.} (A) Molecules are represented through canonical SMILES, AIS-aware tokens, and serialized graph-triplet sequences. (B) \textbf{Training:} the model is trained with token-aware noising that preserves chemically salient tokens, optimizing denoising, consistency, and rounding objectives. (C) \textbf{Inference:} starting from Gaussian noise, iterative denoising and rounding generate either a caption (G2S) or a serialized molecular graph sequence (S2G), which is then deterministically decoded into canonical SMILES.}
    \label{framework}
    \vspace{-4mm}
\end{figure*}

\subsubsection{Diffusion Models: Forward and Reverse Process}

\noindent Denoising diffusion probabilistic models (DDPMs) are generative models that learn a data distribution, often conditioned on some context $\mathbf{c}$, $p(\mathbf{z}_0 \mid \mathbf{c})$. They consist of a fixed forward process and a learned reverse process.

\noindent \textbf{Forward process}: A standard DDPM forward process corrupts clean data $\mathbf{z}_0$ through a Markov chain with noise-schedule coefficients $\{\alpha_t\}_{t=1}^T$ controlling signal decay. This yields the standard closed-form for sampling a noised state $\mathbf{z}_t$ at timestep $t$:
\begin{equation}
    \mathbf{z}_t = \sqrt{\bar{\alpha}_t}\,\mathbf{z}_0 + \sqrt{1 - \bar{\alpha}_t}\,\boldsymbol{\epsilon},
\end{equation}
with $\quad \bar{\alpha}_t = \prod_{s=1}^t \alpha_s \text{ and } \boldsymbol{\epsilon}\sim\mathcal{N}(\mathbf 0,\mathbf I)$. Standard diffusion models typically use a fixed, data-agnostic (isotropic) noise schedule.

\noindent \textbf{Reverse process with Conditional Denoising}: The reverse process learns to recover the clean data $\mathbf{z}_0$ from pure noise $\mathbf{z}_T \sim \mathcal{N}(\mathbf 0,\mathbf I)$. It is defined as a Markov chain $p_\theta(\mathbf{z}_{0:T})$ where each reverse transition $p_\theta(\mathbf{z}_{t-1} \mid \mathbf{z}_t, \mathbf{c})$ is a Gaussian whose mean $\boldsymbol{\mu}_\theta$ and variance $\boldsymbol{\Sigma}_\theta$ are parameterized by a model $\mathcal{M}_\theta(\mathbf{z}_t, t, \mathbf{c})$. The model is trained to predict the mean of the true posterior $q(\mathbf{z}_{t-1} \mid \mathbf{z}_t, \mathbf{z}_0)$. The model parameters $\theta$ are optimized by maximizing the variational lower bound (VLB) on the conditional log-likelihood:
\begin{equation}
\resizebox{\columnwidth}{!}{
$
\begin{aligned}
\mathcal{L}_{\text{vlb}} 
= \mathbb{E}_{q} \Big[
    &\underbrace{-\log p_\theta(\mathbf{z}_0 \mid \mathbf{z}_1, \mathbf{c})}_{\text{Reconstruction } (L_0)} \\
    & + \sum_{t=2}^{T} 
        \underbrace{D_{\mathrm{KL}}\!\left(
            q(\mathbf{z}_{t-1} \mid \mathbf{z}_t, \mathbf{z}_0)
            \,\big\|\,
            p_\theta(\mathbf{z}_{t-1} \mid \mathbf{z}_t, \mathbf{c})
        \right)}_{\text{Denoising Matching } (L_{t-1})} \\
    &+
        \underbrace{D_{\mathrm{KL}}\!\left(
            q(\mathbf{z}_T \mid \mathbf{z}_0)
            \,\big\|\,
            p(\mathbf{z}_T)
        \right)}_{\text{Prior Matching } (L_T)}
\Big].
\end{aligned}
$
}
\end{equation}

\noindent While this formulation is tractable, direct optimization of the full VLB is often unstable.

\subsubsection{Molecules as Graphs:} \label{MAG}A molecule is represented as a graph
$\mathcal{G} = (\mathbf{V}, \mathbf{X}, \mathbf{A}, \mathbf{P})$, 
where $\mathbf{V} = \{1, \dots, n\}$ is the set of atoms (nodes) and $|\mathbf{V}| = n$.
The atom feature matrix $\mathbf{X} \in \mathbb{R}^{n \times a}$ contains
an $a$-dimensional feature vector for each atom.
The adjacency tensor $\mathbf{A} \in \mathbb{R}^{n \times n \times b}$
encodes bond existence and bond types, where
$\mathbf{A}_{ij:} \in \{0,1\}^b$ indicates the type of the bond between atoms $i$ and $j$.
The coordinate matrix $\mathbf{P} \in \mathbb{R}^{n \times 3}$ stores the 3D
positions of atoms.
Here, $a$ is the atom feature dimension and $b$ is the bond type.
Depending on the task, subsets of these components may be used, e.g.,
$(\mathbf{V}, \mathbf{X}, \mathbf{A})$ for 2D graphs or $(\mathbf{V}, \mathbf{X}, \mathbf{P})$
for 3D conformations.
\section{Methodology} \label{Method}



\subsection{Problem Statement}\label{sec:problem}
We use canonical SMILES as the standard molecule representation throughout
\texttt{BiMol-Diff}. From this canonical string view, we derive two related
representations. First, we denote by
$G = \{m_1,\ldots,m_K\}$ the Atoms-in-SMILES (AIS)~\citep{article} token sequence,
which provides a chemistry-aware alternative to character-level SMILES.
Second, for graph-conditioned modeling, we denote by $\tilde{G}$ the
serialized molecular graph obtained by writing the molecule as an
edge-list sequence of atom--bond--atom triplets using special tokens
\texttt{[HEAD]}, \texttt{[REL]}, \texttt{[TAIL]}, and \texttt{[SEP]}:
\[
\tilde{G}
=
\big\langle
\texttt{[HEAD]}~h_i~
\texttt{[REL]}~r_{ij}~
\texttt{[TAIL]}~t_j~
\texttt{[SEP]}
\big\rangle_{(i,j)},
\]
where $(h_i,r_{ij},t_j)$ denotes an atom-bond-atom triple.
Let $S_{\text{text}} = \{s_1,\ldots,s_N\}$ be a textual description.
We study two conditional generation tasks:
(i) \textbf{molecule captioning} $\tilde{G} \rightarrow S_{\text{text}}$, and
(ii) \textbf{molecule generation} $S_{\text{text}} \rightarrow \tilde{G}$.
Accordingly, we learn
$M_\theta : \tilde{G} \rightarrow S_{\text{text}}$ and
$M_\phi : S_{\text{text}} \rightarrow \tilde{G}$ and maximize
\[
\max_{\Theta}
\log p(S_{\text{text}} \mid \tilde{G}; \theta)
+
\log p(\tilde{G} \mid S_{\text{text}}; \phi),
\]
where $\Theta=\{\theta,\phi\}$.
We next describe the molecule-side representation pipeline used to realize these two directions, starting from canonical SMILES and proceeding through AIS-aware tokens and serialized graph sequences.

\subsection{The \texttt{BiMol-Diff} Framework Overview}

We introduce \texttt{BiMol-Diff}, a unified diffusion framework that addresses
molecular graph-to-sequence (G2S) and sequence-to-graph (S2G)
generation under a single conditional modeling view. Given paired data \((\mathcal{A}, \mathcal{B}) \in \{(\mathbf{S_{text}}, \tilde{\mathcal{G}}), (\tilde{\mathcal{G}}, \mathbf{S_{text}})\}\), \texttt{BiMol-Diff} learns the conditional distribution \(p(\mathcal{A}\mid \mathcal{B})\), where \(\mathcal{B}\) provides the conditioning context and \(\mathcal{A}\) is the target modality.

\begin{figure}[h!]
    \centering
    \includegraphics[width=\linewidth]{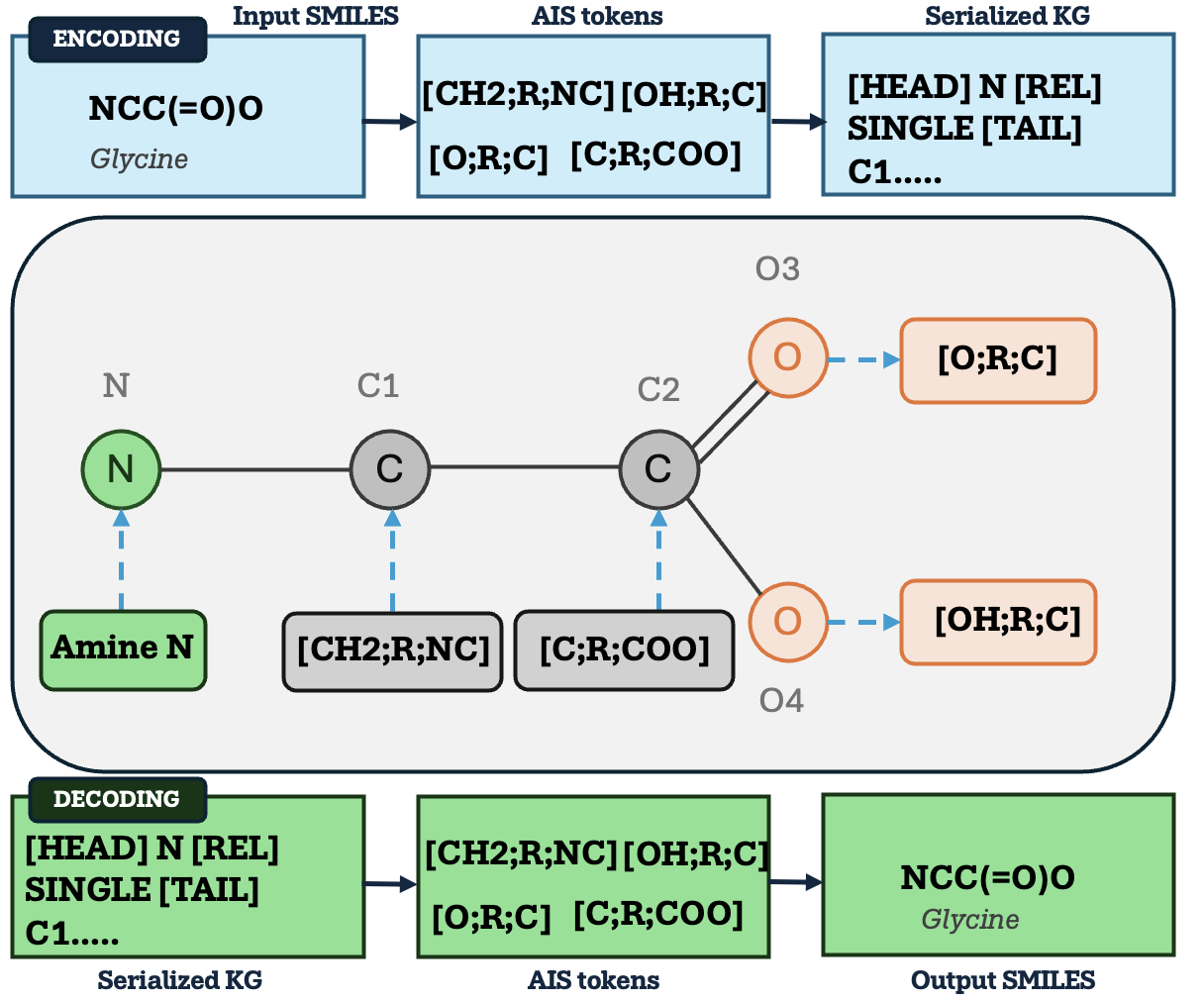}
    \caption{Molecule encoding and decoding in \texttt{BiMol-Diff}.}
    \label{fig:enc_dec}
    \vspace{-3mm}
\end{figure}

\noindent \textbf{Molecule Representation Pipeline.}
We use canonical SMILES as the common molecule interface in both directions and convert it to its Atoms-in-SMILES representation \(G\)~\citep{article}. For graph-conditioned modeling, this molecule-side representation is deterministically mapped to the serialized graph sequence \(\tilde{G}\), where each bond is written as an atom-bond-atom triplet using \texttt{[HEAD]/[REL]/[TAIL]} templates and \texttt{[SEP]} delimiters. Thus, in the G2S direction, \(\tilde{G}\) serves as the conditioning sequence for generating \(\mathbf{S_{text}}\). Conversely, in the S2G direction, \(\mathbf{S_{text}}\) conditions prediction of \(\tilde{G}\), and the predicted \(\tilde{G}\) is deterministically mapped back through the same molecule-side representation pipeline into canonical SMILES for evaluation. Figure~\ref{fig:enc_dec} summarizes this symmetric encoding/decoding view.

\noindent \textbf{Conditional Diffusion.}
For either direction, we embed the target \(\mathcal{A}\) into continuous ``clean'' latents \(\mathbf{z}_0=g_{\Phi}(\mathcal{A})\) and apply a conditional DDPM to obtain \(\mathbf{z}_t\). The reverse processes denoise conditioned on the source modality:
\(\mathcal{M}_\theta(\mathbf{z}_t,t,\tilde{\mathcal{G}})\) for G2S and
\(\mathcal{M}_\phi(\mathbf{z}_t,t,\mathbf{S_{text}})\) for S2G, enabling a unified objective across both directions.

\noindent \textbf{Token-aware Noising.}
Unlike standard data-agnostic schedules that corrupt all tokens uniformly, \texttt{BiMol-Diff} uses a token-aware, token-dependent noising strategy that preserves chemically salient tokens more conservatively. We detail the construction and use of this schedule in Sec~\ref{graph schedule}. This improves
recoverability in both directions while keeping the same diffusion
formulation across the unified framework.



\subsection{Graph Encoding for \texttt{BiMol-Diff}}\label{sec:graph-encoding}
In Section~\ref{sec:problem}, a molecular graph is a set of
relational triplets
\(\tilde{\mathcal{G}} = \{(h_i, r_{ij}, t_j)\}\). To integrate with transformer based encoder-decoder backbone, we
serialize this set of triplets into a single token sequence. Following the
PLM-based G2S convention, each triplet is mapped to a short template
\(\langle \texttt{[HEAD]}~h_i~\texttt{[REL]}~r_{ij}~\texttt{[TAIL]}~t_j \rangle\),
and all such segments are concatenated with a separator token \texttt{[SEP]}:
\begin{equation}
    \resizebox{0.48\textwidth}{!}{$
\begin{aligned}
\tilde{\mathcal{G}} ={}&
\langle\texttt{[HEAD]}~h_{i_1}~\texttt{[REL]}~r_{i_1 j_1}~
\texttt{[TAIL]}~t_{j_1}\rangle~\texttt{[SEP]}~\\\cdots~\texttt{[SEP]}
&\langle\texttt{[HEAD]}~h_{i_M}~\texttt{[REL]}~r_{i_M j_M}~
\texttt{[TAIL]}~t_{j_M}\rangle
\end{aligned}
$}
\end{equation}

\noindent where \(\{(i_m,j_m)\}_{m=1}^M\) enumerates all bonds with \(i_m < j_m\) to avoid
duplicates. This linearized sequence \(\tilde{\mathcal{G}}\) serves
as the conditioning signal to \texttt{BiMol-Diff} in the graph-to-sequence task. Figure~\ref{fig:enc_dec} (top) illustrates this
encoding process.

\noindent \textbf{Example (molecule KG $\rightarrow$ caption).}
Consider the molecule with SMILES string \texttt{CCO} (ethanol). Its molecular
graph has three atoms
$h_1 = \texttt{C}_1$, $h_2 = \texttt{C}_2$, $h_3 = \texttt{O}_3$
and two single bonds: $(1,2)$ and $(2,3)$. The corresponding KG triplets are
$(\texttt{C}_1,\ \texttt{SINGLE},\ \texttt{C}_2)$ and
$(\texttt{C}_2,\ \texttt{SINGLE},\ \texttt{O}_3)$.
Our serialized KG becomes
\[
\resizebox{0.48\textwidth}{!}{$
\begin{aligned}
\tilde{\mathcal{G}} =\;&
\langle \texttt{[HEAD]}~\texttt{C}_1~\texttt{[REL]}~\texttt{SINGLE}~
\texttt{[TAIL]}~\texttt{C}_2 \rangle~\texttt{[SEP]} \\
&\langle \texttt{[HEAD]}~\texttt{C}_2~\texttt{[REL]}~\texttt{SINGLE}~
\texttt{[TAIL]}~\texttt{O}_3 \rangle
\end{aligned}
$}
\]

\noindent In the G2S setting,
\texttt{BiMol-Diff} takes the graph-triplet representation $\tilde{G}$ as input
and generates a natural-language caption $\mathbf{S_{text}}$ describing the
molecule. The reverse realization path used in the S2G direction is
described separately in Section~3.4.

\subsection{Graph Decoding for \texttt{BiMol-Diff}}
In the S2G direction, conditioned on \(\mathbf{S_{text}}\), \texttt{BiMol-Diff}
predicts the serialized molecular graph sequence \(\tilde{G}\). We decode
this sequence by first parsing it into atom-bond-atom triplets
\((h_i, r_{ij}, t_j)\), and then merging these triplets to recover the
corresponding molecular graph. The recovered graph is subsequently mapped
to its AIS token representation \(G\) and canonicalized into a standard
SMILES string for S2G evaluation. Figure~\ref{fig:enc_dec} (bottom)
illustrates this decoding path.

\noindent \textbf{Example (caption $\rightarrow$ molecule KG).}
Consider the textual description corresponding to ethanol. In the S2G
direction, \texttt{BiMol-Diff} predicts the serialized graph sequence:
\begin{equation}
    \tilde{G}
=
\begin{aligned}
&\langle \texttt{[HEAD]}~\texttt{C}_1~\texttt{[REL]}~\texttt{SINGLE}~\texttt{[TAIL]}~\texttt{C}_2 \rangle~\texttt{[SEP]}\\
&\langle \texttt{[HEAD]}~\texttt{C}_2~\texttt{[REL]}~\texttt{SINGLE}~\texttt{[TAIL]}~\texttt{O}_3 \rangle .
\end{aligned}
\end{equation}
These triplets are merged to recover the molecular graph with atoms
\(C_1,C_2,O_3\) and bonds \((1,2)\) and \((2,3)\). The recovered graph
is then mapped to its molecule-side representation and canonicalized into
the SMILES string \texttt{CCO}.

\subsection{Token Aware Noising}
\label{adaptive noise}
\noindent \textbf{Motivation \& Rationale:}\label{motivation}
Standard diffusion models rely on fixed, data-agnostic noising schedules that
apply noise uniformly across all tokens. \\
We argue this is suboptimal for molecular tasks because chemically salient tokens (e.g., functional groups, stereochemistry) typically carry higher information density than structurally redundant or low-information tokens.

\noindent Under uniform noising, chemically salient tokens and
simple syntactic tokens are corrupted at the same rate, which can weaken molecular fidelity. To address this, we introduce a \emph{Token-aware noise
schedule}. We hypothesize that the model’s per-timestep
reconstruction loss $\ell_t^i$ can serve as a useful proxy for token difficulty and importance. By learning a
mapping from $\ell_t^i$ to the corresponding token's noise schedule, we preserve high information tokens while allowing information-redundant tokens to tolerate higher noise levels. This re-parametrizes the denoising path, aiming to improve generation quality in both G2S and S2G tasks.

\noindent \textbf{Noising Schedule}\label{graph schedule}:
Following the rationale discussed in Section~\ref{motivation}, for the G2S setting, we propose a token-dependent noising schedule, parameterized by a vector $\bar{\alpha}_{t, \text{new}}^i \in \mathbb{R}^N$, unlike the uniform noising with baseline cumulative schedule $\bar{\alpha}_t$ (for e.g.: \textit{sqrt}) used by standard diffusion models. This formulation has two stages summarized in Algo~\ref{alg:graph_aware_training}.

\noindent \textbf{Stage 1: Estimating token-wise difficulty.}
For each token $i$ and diffusion step $t=1,\dots,T$, we
define the denoising difficulty as:
\begin{equation}
\label{eq:loss}
\resizebox{0.85\linewidth}{!}{$
    \ell_t^{\,i}
    =
    \mathbb{E}_{\mathbf{z}_t \sim q(\mathbf{z}_t \mid \mathbf{z}_0)}
    \big\|
    \mathcal{M}_\theta(\mathbf{z}_t, t, \tilde{\mathcal{G}})^{(i)}
    - \mathbf{z}_0^{(i)}
    \big\|^2
$}
\end{equation}
Averaging over the training set yields a difficulty profile
$(\ell_1^{\,i},\dots,\ell_T^{\,i})$ for each $i$. Empirically,
$\ell_t^{\,i}$ tends to increase with $t$ (later steps are noisier), but the estimated profile is not strictly monotone. We also compute $\ell_{\min}^{\,i}=\min_t \ell_t^{\,i}$ and $\ell_{\max}^{\,i}=\max_t \ell_t^{\,i}$ to define the difficulty range for token $i$.
In Stage~2 these profiles and their ranges are used to construct a token-wise cumulative schedule, and to obtain a monotonic difficulty profile for each token $i$.

\noindent\textbf{Stage 2: Token-aware schedule.}
Given $(\ell_t^{\,i})_{t=1}^T$ and the baseline cumulative schedule
$(\bar{\alpha}_t)_{t=1}^T$, we construct an adaptive schedule
$(\bar{\alpha}_{t,\mathrm{new}}^{\,i})_{t=1}^T$ for each token $i$. Since this schedule controls noise applied at each step, we want to reallocate noise according to denoising difficulty. Hence, we define a
piecewise-linear map $\Psi_i:[\ell_{\min}^{\,i},\ell_{\max}^{\,i}]\to(0,1)$
that interpolates the baseline schedule as a function of loss:

\begin{equation}
\label{eq:mapping}
\Psi_i(x)
=
\bar{\alpha}_{t-1}
+
\frac{\bar{\alpha}_t - \bar{\alpha}_{t-1}}{\ell_t^{\,i} - \ell_{t-1}^{\,i}}\,
\big(x - \ell_{t-1}^{\,i}\big),
\end{equation}
with $x \in [\ell_{t-1}^{\,i},\, \ell_t^{\,i}),\;\; t=2,\dots,T,$    $\Psi_i(\ell_1^{\,i})=\bar{\alpha}_1$ and $\Psi_i(\ell_T^{\,i})=\bar{\alpha}_T$. In case $\ell_t^{\,i}=\ell_{t-1}^{\,i}$, we add a tiny jitter $\varepsilon$ to avoid division by zero. 
Empirically, $(\ell_t^{\,i})_{t=1}^T$ is not
strictly monotone in $t$, so instead of using its raw values we introduce a new linear ramp in difficulty space:
\begin{equation}
\label{eq:ell_ramp}
\ell_{t}^{\,i,\mathrm{new}}
=
\ell_{\min}^{\,i}
+
\frac{t-1}{T-1}\,\big(\ell_{\max}^{\,i} - \ell_{\min}^{\,i}\big)
\end{equation}
with $t=1,\dots,T$. Substituting this $\ell_{t}^{\,i,\mathrm{new}}$ into $\Psi_i(x)$ we get a new cumulative schedule 
$\bar{\alpha}_{t, \text{new}}^{\,i} = \Psi_i\!\big(\ell_{t}^{\,i,\mathrm{new}}\big)$ for $t=1,\dots,T$. 
We clamp $\bar{\alpha}_{t, \text{new}}^{\,i}$ to $(0,1)$ and apply a non-increasing isotonic projection (refer Appendix~\ref{projection}) over $t$ to obtain the final schedule 
$0 < \bar{\alpha}_{t+1,\mathrm{new}}^{\,i} \le \bar{\alpha}_{t,\mathrm{new}}^{\,i} < 1$ for all $t$.


\begin{algorithm}[t!]
\caption{Token Aware Noise Schedule}
\label{alg:graph_aware_training}
\begin{algorithmic}[1]
\REQUIRE Baseline cumulative schedule $\{\bar{\alpha}_t\}_{t=1}^T$, update interval \texttt{K}
\ENSURE Schedules $\{\bar{\alpha}^{\,i}_{t,\mathrm{new}}\}_{t=1}^T$ for all $i$

\IF{\texttt{train\_step} \% \texttt{K} $== 0$}
  \FOR{\text{all} $i$}
    \STATE Estimate $\{\ell_t^{\,i}\}_{t=1}^T$ via Eq.~\ref{eq:loss}; 
           compute $\{\ell_{\min}^{\,i}, \ell_{\max}^{\,i}\}$.
    \STATE Define piecewise-linear map $\Psi_i$ as in Eq.~(\ref{eq:mapping}).
    \STATE Construct difficulty ramp $\{\ell_{t}^{\,i,\mathrm{new}}\}_{t=1}^T$ via Eq.~(\ref{eq:ell_ramp}).
    \STATE Compute new schedule $\tilde{\alpha}_t^{\,i} = \Psi_i(\ell_{t}^{\,i,\mathrm{new}})$.
    \STATE Clamp $\tilde{\alpha}_t^{\,i}$ to $(0,1)$, apply a non-increasing
           isotonic projection to obtain 
           $\{\bar{\alpha}^{\,i}_{t,\mathrm{new}}\}_{t=1}^T$.
  \ENDFOR
\ENDIF

\RETURN $\{\bar{\alpha}^{\,i}_{t,\mathrm{new}}\}$ for all tokens $i$ and steps $t$.
\end{algorithmic}
\end{algorithm}
\vspace{-4mm}

\subsection{Model Training and Inference}
\begin{table*}[t!]
    \centering
    \tiny
    \resizebox{0.8\textwidth}{!}{%
    \begin{tabular}{l c c c c c c}
        \toprule
        \textbf{Method} & \textbf{\#P} & \textbf{B} & \textbf{CrF++} & \textbf{M} & \textbf{B-F1} & \textbf{MVE} \\
        \midrule
        
        \multicolumn{7}{l}{\textbf{Autoregressive Baselines}} \\
        MolT5-Base~\citep{edwards-etal-2022-translation} & 220M & 0.452 & 0.651 & 0.510 & 0.681 & 0.852 \\
        Text+Chem T5~\citep{christofidellis2023textchemt5} & 223M & \underline{0.542} & \underline{0.701} & 0.648 & 0.728 & 0.866 \\
        MolCA~\citep{liu2024molcamoleculargraphlanguagemodeling} & 110M & 0.531 & 0.665 & \textbf{0.651} & 0.709 & 0.815 \\
        GitMol~\citep{Liu_2024} & 700M & 0.475 & 0.680 & 0.532 & 0.751 & 0.875 \\
        GraphT5~\citep{kim2025grapht5unifiedmoleculargraphlanguage} & 272M & 0.481 & 0.692 & 0.545 & \underline{0.810} & \underline{0.913} \\

        \midrule
        
        \multicolumn{7}{l}{\textbf{Diffusion Baselines}} \\
        Diffusion-LM~\citep{li2022diffusionlmimprovescontrollabletext} & 91M & 0.512 & 0.702 & 0.602 & 0.783 & 0.861 \\
        DiffuSeq~\citep{gong2023diffuseqsequencesequencetext} & 91M & 0.532 & 0.708 & 0.601 & 0.812 & 0.887 \\
        TGM-DLM~\citep{gong2024tgmdlm} & 125M & 0.467 & 0.689 & 0.589 & 0.779 & 0.856 \\

        \midrule
        
        \rowcolor{white!70!yellow}
        \rowcolor{white!70!yellow}
        \texttt{BiMol-Diff} (ours) & 63M & \textbf{0.567} & \textbf{0.734} & 0.626 & \textbf{0.843} & \textbf{0.925} \\

        \rowcolor{white!70!gray}
        \%Gain (vs. Best AR) & \textbf{x3.5$\downarrow$} & \textbf{+4.6\%} & \textbf{+4.7\%} & -3.8\% & \textbf{+4.1\%} & \textbf{+1.3\%} \\
        \rowcolor{white!70!gray}
        \%Gain (vs. Best Diff) & \textbf{x1.4$\downarrow$} & \textbf{+6.6\%} & \textbf{+3.7\%} & \textbf{+4.2\%} & \textbf{+3.8\%} & \textbf{+4.3\%} \\
        \bottomrule
    \end{tabular}%
    }
    \caption{Molecule captioning performance on the M3-20M dataset. Benchmarking \texttt{BiMol-Diff} against SoTA Autoregressive (AR) and Diffusion models. \texttt{BiMol-Diff} achieves state-of-the-art results, surpassing the best diffusion and AR baselines by substantial margins across all metrics while using fewer parameters.}
    \label{molecule captioning}
    \vspace{-3mm}
\end{table*}

\noindent \textbf{Training}: Our training objective is derived from the Variational Lower Bound (VLB) (Eq. 2) presented in the preliminaries. While the full VLB optimization can be unstable \citep{ho2020denoisingdiffusionprobabilisticmodels}, 
a common simplification is to train the model $\mathcal{M}_\theta$ to predict the added noise $\boldsymbol{\epsilon}$. However, our framework adopts an alternative $\mathbf{z}_0$-prediction reparameterization, which trains the model to directly predict the clean data $\mathbf{z}_0$ at every timestep $t$.
A critical component of this objective is the rounding term $L_0 = -\log \tilde{p}_\Phi(\mathbf{S_{text}} \mid \mathbf{z}_0)$, which handles the final step of converting the continuous latent variable $\mathbf{z}_0$ back into discrete tokens $\mathbf{S_{text}}$. We define this as a trainable rounding distribution: $\tilde{p}_\Phi(\mathbf{S_{text}} \mid \mathbf{z}_0) = \prod_{i=1}^{N} \tilde{p}_\Phi(s_i \mid \mathbf{z}_{0,i})$,
where each token $s_i$ is sampled from a softmax distribution over the vocabulary, using logits derived from the corresponding output embedding $\mathbf{z}_{0,i}$.
By combining this rounding term (for $t=0$) with the denoising matching terms (for $t>1$) using our $\mathbf{z}_0$-prediction reparameterization, we arrive at our final, composite objective. (The full derivation from the VLB is in App \ref{sec:derivation_appendix}).

\begin{equation}
\label{z0-pred}
\resizebox{0.85\columnwidth}{!}{
$
\begin{aligned}
\mathcal{L}_{\text{e2e-simple}}(\mathbf{S_{text}}) = \mathbb{E}_{q} 
    \Bigl[ \sum_{t=2}^{T} \underbrace{\|\mathcal{M}_\theta(\mathbf{z}_t, t, \tilde{\mathcal{G}}) - \mathbf{z}_0 \|^2}_{\text{Denoising}} + \\\underbrace{\|\,g_\Phi(\mathbf{S_{text}}) - \mathcal{M}_\theta(\mathbf{z}_1, 1, \tilde{\mathcal{G}})\|^2}_{\text{Consistency}} - \underbrace{\log \tilde{p}_\Phi(\mathbf{S_{text}} \mid \mathbf{z}_0)}_{\text{Rounding}} \Bigr]
\end{aligned}
$
}
\end{equation}
The same objective is used in the S2G direction, with the target discrete sequence replaced by \(\tilde{G}\) and the conditioning sequence replaced by \(S_{\text{text}}\).
This objective directly optimizes the most critical parts of the process: the denoising accuracy across all steps (Denoising), the consistency of the first denoising step with the true data embedding (Consistency), and the quality of the final conversion to discrete tokens (Rounding)\label{rounding}. 

\noindent \textbf{Inference-time schedule:}
The $\mathbf{z}_0$-prediction objective (Eq:~\ref{z0-pred}) trains the denoiser to explicitly predict the clean latent at each step.
At inference, given $\mathbf{z}_t$ and conditioning context $\mathbf{c}$, we compute
$\hat{\mathbf{z}}_0 = \mathcal{M}_\theta(\mathbf{z}_t, t, \mathbf{c})$
(with $\mathbf{c}=\tilde{\mathcal{G}}$ for G2S and $\mathbf{c}=\mathbf{S_{text}}$ for S2G).
We then use our learned token-wise cumulative schedule $\bar{\alpha}^{\,i}_{t,\mathrm{new}}$ in the reverse sampling update, where $\boldsymbol{\epsilon}\sim\mathcal{N}(\mathbf{0},\mathbf{I})$.
To mitigate rounding errors (refer section~\ref{rounding}), we apply the clamping trick, which forces the predicted vector to commit to a word for intermediate diffusion steps:
\begin{equation}
\label{eq:inference_schedule}
\resizebox{0.85\columnwidth}{!}{$
\begin{aligned}
\mathbf{z}^{(i)}_{t-1}
&=
\sqrt{\bar{\alpha}^{\,i}_{t-1,\mathrm{new}}}\;\mathrm{Clamp}\!\Big(\mathcal{M}_\theta(\mathbf{z}_t, t, \mathbf{c})^{(i)}\Big) \\
&\quad+\;
\sqrt{1-\bar{\alpha}^{\,i}_{t-1,\mathrm{new}}}\;\boldsymbol{\epsilon}^{(i)},
\end{aligned}
$}
\end{equation}
\noindent where $\mathrm{Clamp}(\cdot)$ denotes nearest-neighbor projection onto the embedding table.
This strategy ensures that the diffusion trajectory remains grounded in valid token embeddings during sampling.

\begin{table*}[t!]
    \centering
    \tiny
    \resizebox{0.8\textwidth}{!}{%
    \begin{tabular}{l c c c c c c c}
        \toprule
        \textbf{Method} & \textbf{Type} & \textbf{\#P} & \textbf{Exact} & \textbf{MACCS} & \textbf{RDK} & \textbf{Morgan} & \textbf{Valid} \\
        \midrule
        \multicolumn{8}{l}{\textbf{Text-Based Autoregressive (SMILES)}} \\
        T5-Base~\citep{JMLR:v21:20-074} & AR & 248M & 0.069 & 0.731 & 0.605 & 0.545 & 0.660 \\
        MolT5-Base~\citep{edwards-etal-2022-translation} & AR & 248M & 0.081 & 0.721 & 0.588 & 0.529 & 0.772 \\
        MolXPT~\citep{liu-etal-2023-molxpt} & AR & 350M & 0.215 & \textbf{0.859} & \textbf{0.757} & \textbf{0.667} & \textbf{0.983} \\
        T5 Enc + MolXPT MLP~\citep{Deng_2025} & Adapter & 111M & \textbf{0.227} & 0.820 & 0.713 & 0.636 & 0.980 \\
        \midrule
        \multicolumn{8}{l}{\textbf{Text-Based Autoregressive (SELFIES)}} \\
        SciBERT + MolGen~\citep{Deng_2025} & Adapter & 317M & 0.083 & 0.680 & 0.526 & 0.422 & 0.995 \\
        Galactica-1.3B + MolGen~\citep{fang2024domainagnosticmoleculargenerationchemical} & Adapter & 1.5B & 0.091 & 0.706 & 0.560 & 0.454 & \underline{0.995} \\
        T5 Enc + MolGen~\citep{Deng_2025} & Adapter & 317M & 0.165 & 0.758 & 0.616 & 0.527 & 0.995 \\
        \midrule
        \multicolumn{8}{l}{\textbf{Graph \& Diffusion Baselines}} \\
        DiGress (sim. guidance)~\citep{vignac2023digressdiscretedenoisingdiffusion} & Diff & 289M & 0.014 & 0.577 & 0.389 & 0.288 & 0.854 \\
        3M-Diffusion~\citep{zhu20243mdiffusionlatentmultimodaldiffusion} & Diff & 162M & 0.005 & 0.548 & 0.370 & 0.273 & \textbf{1.000} \\
        Graph-DiT~\citep{liu2024graphdit} & Diff & 162M & 0.000 & 0.374 & 0.269 & 0.159 & 0.909 \\
        UTGDiff (w/o pretrain)~\citep{xiang2025instructionbasedmoleculargraphgeneration} & Diff & 125M & \underline{0.227} & \underline{0.867} & \underline{0.763} & \underline{0.695} & 0.856 \\
        \midrule
        
        \rowcolor{white!70!yellow}
        \texttt{BiMol-Diff} (ours) & Diff & 180M & \textbf{0.262} & \textbf{0.894} & \textbf{0.791} & \textbf{0.762} & 0.901 \\
        
        \rowcolor{white!70!gray}
        \%Gain (vs. Best AR) & -- & \textbf{x1.9$\downarrow$} & \textbf{+15.4\%} & \textbf{+4.1\%} & \textbf{+4.5\%} & \textbf{+14.2\%} & -8.3\% \\

        \rowcolor{white!70!gray}
        \%Gain (vs. Best Diff) & -- & \textbf{x1$\downarrow$} & \textbf{+15.4\%} & \textbf{+3.1\%} & \textbf{+3.7\%} & \textbf{+9.6\%} & \textbf{+5.3\%} \\
        \bottomrule
    \end{tabular}%
    }
    \caption{Molecule generation performance on the ChEBI-20 test set. \texttt{BiMol-Diff} outperforms state-of-the-art Autoregressive and Diffusion baselines in structure reconstruction (Exact Match) and fingerprint similarity, validating the effectiveness of token-aware denoising.}
    \label{molecule generation}
    \vspace{-4mm}
\end{table*}

\section{Experiments} \label{experiments}

\subsection{Experimental Setup}
\noindent \textbf{Model Architecture.} We implement \texttt{BiMol-Diff} using two distinct encoder-decoder Transformer models tailored for each task utilizing GeLU activations. Crucially, both tasks are trained for 200,000 steps using our proposed \textit{Token-aware noise schedule} (refer~\S\ref{adaptive noise}) over $T=2000$ diffusion steps. We adapt the architecture for each task as follows: for Molecule Generation (S2G), we employ a 6-encoder/12-decoder configuration ($d_{model}=1024$; $\approx 180$M parameters). The encoder processes
frozen SciBERT embeddings, while the decoder predicts the serialized
molecular graph sequence \(\tilde{G}\). This predicted graph sequence is
then deterministically mapped back into a canonical SMILES string for
evaluation. This variant is trained with a batch size of 64 and a learning rate of $5 \times 10^{-5}$. Conversely, for Molecule Captioning (G2S), we utilize a 6-encoder/9-decoder configuration ($d_{model}=512$; $\approx 63$M parameters) with a peak learning rate of $10^{-4}$. This model encodes serialized molecular structure using the AIS-based SMILES vocabulary augmented with learnable special tokens (\texttt{[HEAD]}, \texttt{[REL]}, \texttt{[TAIL]}, \texttt{[SEP]}), while the decoder generates captions utilizing the \texttt{bert-base-uncased} vocabulary~\cite{devlin-etal-2019-bert}.

\noindent \textbf{Datasets.} We evaluate our framework on two distinct datasets tailored to each task. For Molecule Generation (S2G), we utilize the ChEBI-20~\citep{edwards-etal-2022-translation} benchmark, comprising 33,010 molecules with SMILES strings restricted to a maximum length of 256. This dataset is partitioned into an 80\%/10\%/10\% split for training, validation, and testing, respectively. For Molecule Captioning (G2S), we employ a filtered subset of the M3-20M dataset~\cite{guo2025m320mlargescalemultimodalmolecule}, containing approximately 360,000 SMILES--description pairs. This subset, also divided into an 80\%/10\%/10\% split, pairs molecular graphs with diverse textual descriptions derived from curated scientific literature or synthesized via GPT-3.5.

\noindent \textbf{Baselines.} We compare \texttt{BiMol-Diff} against state-of-the-art (SoTA) autoregressive and diffusion-based methods for each task. 
For Molecule Generation (S2G), we benchmark against MolXPT~\citep{liu-etal-2023-molxpt} as a competitive autoregressive SMILES generator, and leading diffusion baselines including 3M-Diffusion~\citep{zhu20243mdiffusionlatentmultimodaldiffusion} and UTGDiff~\citep{xiang2025instructionbasedmoleculargraphgeneration}.
For Molecule Captioning (G2S), we compare against MolT5-Base~\citep{edwards-etal-2022-translation} as a standard SMILES$\rightarrow$text sequence-to-sequence baseline, GraphT5~\citep{kim2025grapht5unifiedmoleculargraphlanguage} as a recent graph--text model, and DiffuSeq~\citep{gong2023diffuseqsequencesequencetext} as a representative diffusion-based sequence generator.

\noindent \textbf{Evaluation Metrics.} For Molecule Captioning (G2S), we evaluate the quality of generated text using both surface-level and semantic measures. We report standard $n$-gram overlap metrics including BLEU \citep{papineni2002bleu}, METEOR \citep{banerjee2005meteor}, and ChrF++ \citep{popovic-2015-chrf}, alongside embedding-based metrics such as BERTScore-F1 \citep{zhang2020bertscoreevaluatingtextgeneration} and MAUVE \citep{pillutla2023mauvescoresgenerativemodels} to assess semantic alignment and distributional divergence. For Molecule Generation (S2G), we focus on structural fidelity and chemical plausibility. We report Exact Match accuracy to measure precise reconstruction of the ground-truth molecule. To assess the capture of chemical substructures, we compute structural overlap using Tanimoto similarity on MACCS, RDKit, and Morgan fingerprints. Finally, we report chemical validity to ensure the generated SMILES strings correspond to valid molecular graphs.

\subsection{Experimental Results}
\noindent \textbf{Molecule Captioning (G2S):} Table~\ref{molecule captioning} reports results on M3-20M. \texttt{BiMol-Diff} achieves the best overall performance on BLEU and ChrF++, reaching \textbf{0.567} BLEU and \textbf{0.734} ChrF++, which improves over the best performing autoregressive baseline (Text+Chem T5) by \textbf{4.6\%} and \textbf{4.7\%}, respectively. On semantic metrics, our model attains \textbf{0.843} BERTScore-F1 and \textbf{0.925} MAUVE, outperforming graph-aware baselines such as GraphT5, and consistently improving over diffusion models (e.g., DiffuSeq) across all reported metrics. While MolCA achieves the highest METEOR score, \texttt{BiMol-Diff} remains competitive (\textbf{0.626}) and offers a strong overall trade-off across surface and embedding-based measures, supporting the benefit of token-aware noising.

\noindent \textbf{Molecule Generation (S2G):} Table~\ref{molecule generation} summarizes results on ChEBI-20. \texttt{BiMol-Diff} attains an Exact Match of \textbf{0.262}, improving over both the best autoregressive adapter baseline (T5 Enc + MolXPT MLP) and the strongest diffusion baseline (UTGDiff), each at 0.227 (\textbf{+15.4\%} relative). This gain is reflected in fingerprint similarities, achieving \textbf{0.894} (MACCS), \textbf{0.791} (RDKit), and \textbf{0.762} (Morgan), including a \textbf{14.2\%} improvement in Morgan similarity over the best autoregressive baseline. Finally, \texttt{BiMol-Diff} maintains high chemical validity (\textbf{0.901}) while prioritizing reconstruction accuracy, whereas some baselines attain higher validity with weaker structural recovery.
\subsection{Ablation}
\begin{table*}[t!]
    \centering
    \small
    \setlength{\tabcolsep}{10pt}
    \resizebox{0.8\textwidth}{!}{%
    \begin{tabular}{l l c c c}
        \toprule
        \textbf{Noise Schedule} & \textbf{Mapping / Tokenizer} & \textbf{B} & \textbf{CrF++} & \textbf{M} \\
        \midrule
        
        \multicolumn{5}{l}{\textbf{\textit{Impact of Noise Schedule \& Mapping Function}}} \\
        Uniform & Sqrt Schedule & 0.495 & 0.682 & 0.531 \\
        Token-Aware & Cosine Mapping & 0.548 & 0.720 & 0.595 \\
       
        \rowcolor{yellow!25}
        Token-Aware (Ours) & Linear Mapping & \textbf{0.567} & \textbf{0.734} & \textbf{0.626} \\
        
        \midrule
        
        \multicolumn{5}{l}{\textbf{\textit{Impact of Molecular Tokenization}}} \\
        Token-Aware (Linear) & Regex-based & 0.558 & 0.714 & 0.606 \\
        Token-Aware (Linear) & Atom-level & 0.563 & 0.722 & 0.613 \\
        \rowcolor{yellow!25}
        Token-Aware (Linear) & Atoms-in-SMILES (AIS) & \textbf{0.567} & \textbf{0.734} & \textbf{0.626} \\
        \bottomrule
    \end{tabular}}
    \caption{Ablation study on the M3-20M dataset. We analyze the impact of different (\textbf{Top}) Noise Schedules \& Mapping Functions and (\textbf{Bottom}) Tokenization strategies on molecule captioning performance. \texttt{BiMol-Diff} (Token-Aware + Linear) with Atoms-in-SMILES (AIS) tokenization yields the best results.}
    \label{tab:ablation_study}
    \vspace{-4mm}
\end{table*}

\noindent We analyze three aspects of \texttt{BiMol-Diff}: (i) the proposed token-aware noising strategy and its mapping function, (ii) the effect of molecular tokenization, and (iii) decoding efficiency of \texttt{BiMol-Diff}.

\begin{figure}[ht!]
    \centering
    \includegraphics[width=\linewidth]{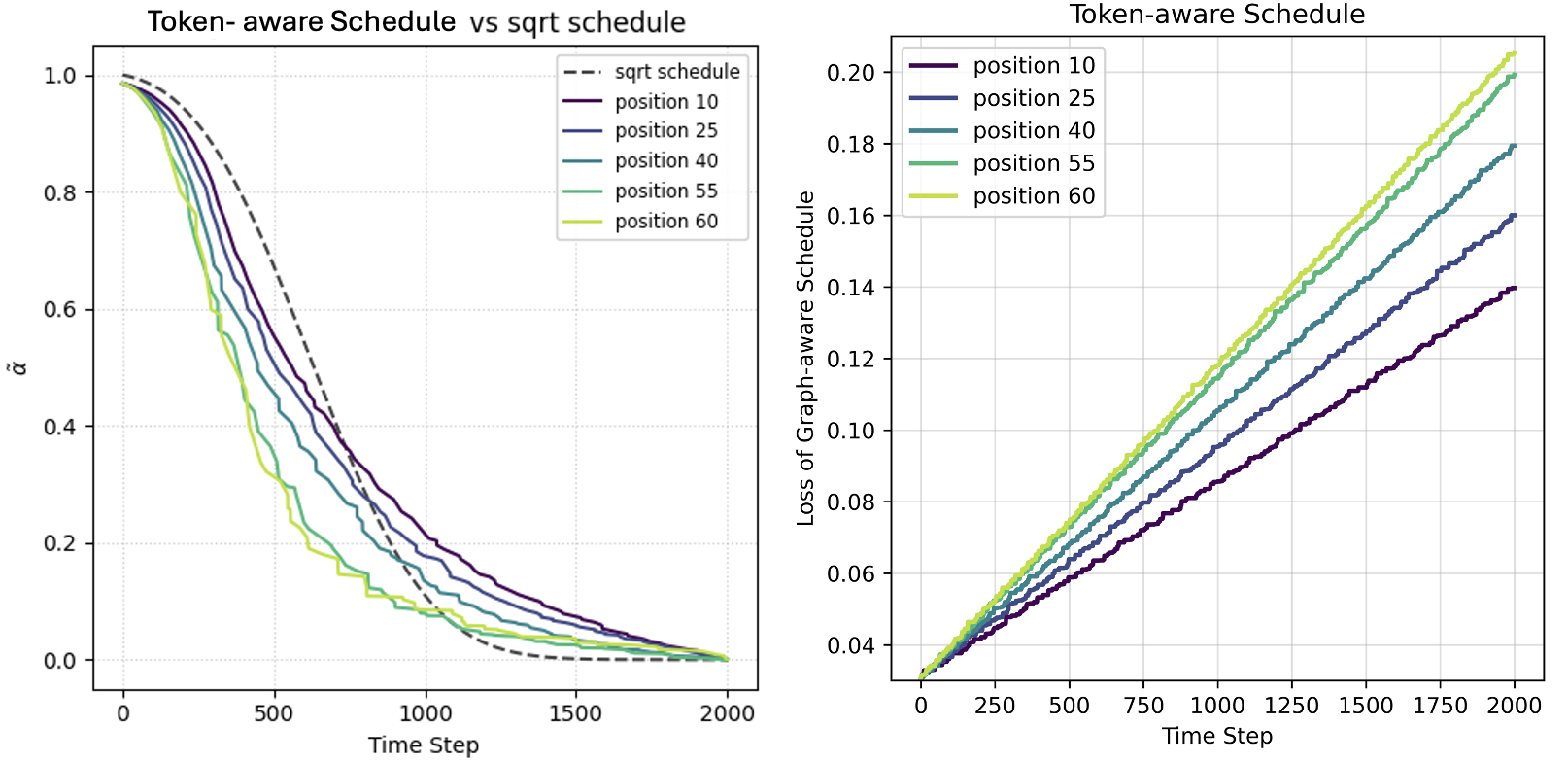}
    \caption{Token-aware noising vs.\ uniform sqrt schedule (captioning). \emph{Left:} the learned token-wise schedules apply non-uniform corruption across positions compared to a global sqrt schedule. \emph{Right:} token-wise difficulty profiles are used to construct schedules that apply more conservative corruption to harder-to-recover tokens.}
    \label{noise schedule}
\end{figure}

\noindent \textbf{(i) Impact of noise schedule and choice of mapping function.}
Table~\ref{tab:ablation_study} (top) shows that replacing a uniform corruption schedule (\textit{sqrt}) with our token-aware schedule yields a substantial improvement (0.495$\rightarrow$0.567 BLEU; 0.682$\rightarrow$0.734 ChrF++; 0.531$\rightarrow$0.626 METEOR). Among mapping choices, the linear mapping performs best, indicating that explicitly allocating lower corruption to harder-to-recover tokens is beneficial for caption recovery. Figure~\ref{noise schedule} visualizes how token-aware schedules differ from a global \textit{sqrt} schedule. Instead of applying the same noise profile to all positions, we learn per-token schedules that modulate the corruption level according to token difficulty: positions with higher denoising loss receive more conservative corruption (lower noise at intermediate timesteps), while easier positions can be noised more aggressively. This selective corruption better preserves content during the forward process and improves recoverability in the reverse process, which aligns with the gains observed in the main results (Table~\ref{molecule captioning},~\ref{molecule generation}).
\vspace{-3mm}\\
\noindent \textbf{(ii) Impact of molecular tokenization.}
Table~\ref{tab:ablation_study} (bottom) applies the best noising configuration (token-aware + linear) with different molecular tokenizers. Our current framework uses an AIS-based SMILES tokenizer for a stable, deterministic segmentation that preserves SMILES syntax (e.g., bracketed atoms and ring/branch symbols) and enables straightforward serialization, the tokenizer is plug-and-play and can be swapped without changing the denoising objective or architecture. We observe a clear monotonic trend: regex-based SMILES tokenization underperforms atom-level tokens, while Atoms-in-SMILES (AIS) yields the strongest performance across all reported metrics.



\noindent \textbf{(iii) Decoding Efficiency.} Given the iterative nature of diffusion models, we examine the trade-off between generation quality and inference latency. Reducing the number of reverse steps leads to substantial speedups, albeit with a degradation in output fidelity. To focus on modeling performance, all primary results are reported with $T{=}2000$ steps. Table~\ref{tab:efficiency_g2s_merged} presents the quality-latency trade-off for Graph-to-Sequence tasks. Within the diffusion family, \texttt{BiMol-Diff} exhibits markedly improved efficiency over DiffuSeq. At $T{=}2000$, it achieves a BLEU score of 0.567 in 89 seconds per batch, yielding a $3.6\times$ speedup over DiffuSeq (317 seconds), while also improving generation quality.
\begin{table}[ht!]
\centering
\small
\setlength{\tabcolsep}{4pt}
\renewcommand{\arraystretch}{1.2}

\resizebox{\columnwidth}{!}{%
\begin{tabular}{l c c c c c}
\toprule
\textbf{Method} &
\textbf{Params} &
\textbf{Steps} &
\textbf{Time (s)} &
\textbf{Speedup} &
\textbf{BLEU $\uparrow$} \\
\midrule
\multicolumn{6}{l}{\textit{\textbf{Autoregressive Baselines}}} \\
MolT5-Base \citep{edwards-etal-2022-translation} & 220M & -- & $<$5s & -- & 0.452 \\
Text+Chem T5 \citep{christofidellis2023textchemt5} & 223M  & -- & $<$5s & -- & 0.542 \\
\midrule
\multicolumn{6}{l}{\textit{\textbf{Diffusion Baselines}}} \\
DiffuSeq \citep{gong2023diffuseqsequencesequencetext}     & 91M  & 2000 & 317s & $1.0\times$ (Ref) & 0.532 \\
\midrule
\multicolumn{6}{l}{\textit{\textbf{Ours}}} \\
\textbf{BiMol-Diff} & 63M & 2000 & 89s & $3.6\times$ & \textbf{0.567} \\
\textbf{BiMol-Diff} & 63M & 1000 & 45s & $7.0\times$ & 0.551 \\
\textbf{BiMol-Diff} & 63M & 500  & 23s & $13.8\times$ & 0.365 \\
\textbf{BiMol-Diff} & 63M & 100  & 5s  & $63.4\times$ & 0.312 \\
\bottomrule
\end{tabular}%
}
\caption{Inference efficiency comparison on the G2S task. We compare \texttt{BiMol-Diff} against the best Autoregressive (AR) and Diffusion baselines from Table \ref{molecule captioning}. Time is measured as total inference time for a batch size of 50 on a single NVIDIA V100 GPU.}
\label{tab:efficiency_g2s_merged}

\end{table}
\noindent When compared to Autoregressive (AR) baselines, our model shows a clear operational boundary. In the high-quality output range (1000--2000 steps), \texttt{BiMol-Diff} outperforms the strongest AR baseline, Text+Chem T5 (0.542 BLEU), confirming that the extra computational time yields better text generation. However, the model struggles when pushed for extreme speed. As we reduce the inference steps below 1000, we observe a sharp drop in performance. At 500 steps, although the inference time improves to 23 seconds (13.8$\times$ speedup), the BLEU score falls to 0.365, which is lower than even the base MolT5 model (0.452). This indicates that \texttt{BiMol-Diff} requires a minimum threshold of denoising steps (approximately 1000) to maintain its advantage. Our model is ideal for high-precision generation and simple AR models remain the better choice for low-latency applications.
\section{Conclusion} \label{conclusion}
\vspace{-1mm}
We introduced \texttt{BiMol-Diff}, a unified diffusion framework addressing the bidirectional tasks of molecular graph generation and textual captioning. By moving beyond standard data-agnostic corruption to a \textit{token-aware noise schedule}, our approach explicitly preserves chemically salient substructures during the diffusion process. Empirical validation on M3-20M and ChEBI-20 confirms that this strategy yields significant gains over state-of-the-art autoregressive and diffusion baselines, particularly in structural fidelity (S2G) and semantic alignment (G2S). These findings demonstrate that token-aware denoising is essential for high-fidelity, controllable molecular language modeling, offering a robust alternative to autoregressive approaches in scientific domains.

\section{Limitations}
First, \texttt{BiMol-Diff} relies on linearized sequence representations and standard attention, lacking the explicit structural inductive bias inherent to graphs, which may limit data efficiency for complex topologies. Second, as with most diffusion models, our approach incurs higher computational costs during inference compared to autoregressive baselines due to the iterative denoising process, highlighting the need for future work on accelerated sampling.
\bibliography{custom}
\appendix

\section{Appendix}\label{sec:App}

\subsection{Related Works}\label{related work}
Table~\ref{relatedwork_table} positions \texttt{BiMol-Diff} relative to recent state-of-the-art molecule--text generation frameworks. While autoregressive (AR) baselines, such as MolT5 and UniMoT, have successfully demonstrated bidirectional capabilities, they often suffer from the standard limitations of sequential decoding (e.g., exposure bias). Conversely, existing diffusion-based approaches like TGM-DLM and UTGDiff offer non-autoregressive benefits but are predominantly unidirectional, focusing almost exclusively on the Text$\to$Graph modality. \texttt{BiMol-Diff} bridges this gap as a unified diffusion framework capable of both Graph$\to$Text and Text$\to$Graph generation. Furthermore, we distinguish our approach by integrating a \textit{token-aware mechanism}, a feature notably absent in prior diffusion baselines—which adapts the noising schedule to prioritize chemically significant tokens, similar to the strategy employed by the AR-based ChemT5.

\begin{table*}[t!]
\centering
\small
\setlength{\tabcolsep}{5pt}
\renewcommand{\arraystretch}{1.18}

\resizebox{\textwidth}{!}{%
\begin{tabular}{l l c c c c c}
\toprule
\textbf{Method} &
\textbf{Generation family} &
\textbf{Text-conditioned} &
\textbf{Graph$\rightarrow$Text} &
\textbf{Text$\rightarrow$Graph} &
\textbf{Bidirectional} &
\textbf{Token-aware training} \\
\midrule
\multicolumn{7}{l}{\textit{\textbf{Autoregressive (AR)}}} \\
\midrule
MolT5 \citep{edwards-etal-2022-translation} &
Seq2Seq PLM (T5) &
\checkmark & \checkmark & \checkmark & \checkmark & --- \\
Text+ChemT5 \citep{christofidellis2023textchemt5} &
Seq2Seq PLM (T5) &
\checkmark & \checkmark & \checkmark & \checkmark & --- \\
CAMT5 \citep{kim2025camt5} &
Seq2Seq PLM (T5) &
\checkmark & --- & \checkmark & --- & \checkmark \\
UniMoT \citep{guo2025unimotunifiedmoleculetextlanguage} &
LLM-based multitask &
\checkmark & \checkmark & \checkmark & \checkmark & --- \\
LDMol \citep{chang2025ldmoltexttomoleculediffusionmodel} &
PLM-based generator &
\checkmark & --- & \checkmark & --- & --- \\
MolCA \citep{liu2024molcamoleculargraphlanguagemodeling} &
Encoder--decoder (graph-aware) &
--- & \checkmark & --- & --- & --- \\
Mol2Lang-VLM \citep{tran-etal-2024-mol2lang} &
Multimodal PLM &
--- & \checkmark & --- & --- & --- \\
\midrule
\multicolumn{7}{l}{\textit{\textbf{Non-autoregressive (Diffusion)}}} \\
\midrule
TGM-DLM \citep{gong2024tgmdlm} &
Diffusion-LM &
\checkmark & --- & \checkmark & --- & --- \\
3M-Diffusion \citep{zhu20243mdiffusionlatentmultimodaldiffusion} &
Latent diffusion (graph/seq) &
\checkmark & --- & \checkmark & --- & --- \\
UTGDiff \citep{xiang2025instructionbasedmoleculargraphgeneration} &
Graph diffusion &
\checkmark & --- & \checkmark & --- & --- \\
\rowcolor{yellow!18}
\texttt{BiMol-Diff} (ours) &
Conditional diffusion &
\checkmark & \checkmark & \checkmark & \checkmark & \checkmark \\
\bottomrule
\end{tabular}%
}
\caption{Feature comparison of \texttt{BiMol-Diff} against representative autoregressive (AR) and diffusion baselines. (\checkmark = supported; --- = not supported/reported). Token-aware mechanism denotes chemistry-aware tokenization/weighting in AR models and token-wise noising schedules in diffusion models. Unlike prior diffusion works which are typically unidirectional, \texttt{BiMol-Diff} enables fully bidirectional generation.}
\label{relatedwork_table}
\end{table*}

\subsection{Derivation}\label{sec:derivation_appendix}

\texttt{BiMol-Diff} builds on the standard diffusion framework, which trades the flexibility of expressive generative models (e.g., GANs, VAEs, flow models) for the tractability of likelihood-based training in a continuous latent space $\mathbf{z}$. The overall goal is to minimize the negative log-likelihood
\begin{equation}
    \mathbb{E}_{\mathbf{z}_0,\mathbf{c}}\!\bigl[-\log p_\theta(\mathbf{z}_0 \mid \mathbf{c})\bigr],
\end{equation}
which is upper-bounded by the Variational Lower Bound (VLB).

\subsubsection{Forward and Reverse Processes}

The forward Markov chain is defined as $ q(\mathbf{z}_{1:T} \mid \mathbf{z}_0)
    = \prod_{t=1}^T q(\mathbf{z}_t \mid \mathbf{z}_{t-1}),$
where each transition is Gaussian:
\begin{equation}
    q(\mathbf{z}_t \mid \mathbf{z}_{t-1})
    = \mathcal{N}\!\Bigl(
        \mathbf{z}_t \,\big|\,
        \sqrt{1-\beta_t}\,\mathbf{z}_{t-1},\,
        \beta_t\,\mathbf{I}
    \Bigr).
\end{equation}
Let $\alpha_t = 1-\beta_t$ and $\bar{\alpha}_t = \prod_{i=1}^t \alpha_i$.  
By induction, the marginal at time $t$ satisfies:
\begin{equation}
    \mathbf{z}_t
    = \sqrt{\bar{\alpha}_t}\,\mathbf{z}_0
    + \sqrt{1-\bar{\alpha}_t}\,\boldsymbol{\epsilon},
    \qquad
    \boldsymbol{\epsilon}\sim\mathcal{N}(\mathbf{0},\mathbf{I}),
\end{equation}
so that $ q(\mathbf{z}_t \mid \mathbf{z}_0)
    = \mathcal{N}\!\Bigl(
        \sqrt{\bar{\alpha}_t}\,\mathbf{z}_0,\,
        (1-\bar{\alpha}_t)\mathbf{I}
    \Bigr).$
We used the \textit{sqrt} schedule as the baseline schedule used in DiffusionLM \cite{li2022diffusionlmimprovescontrollabletext}, namely \(\bar\alpha_t=1-\sqrt{t/T+s}\) with small \(s>0\).
The reverse denoising process then learns

\begin{equation}
\begin{aligned}
p_\theta(\mathbf{z}_{0:T})
&= p(\mathbf{z}_T)\prod_{t=1}^T p_\theta(\mathbf{z}_{t-1}\mid \mathbf{z}_t), \\
p_\theta(\mathbf{z}_{t-1}\mid \mathbf{z}_t)
&= \mathcal{N}\!\Bigl(\boldsymbol{\mu}_\theta(\mathbf{z}_t,t),\,\boldsymbol{\sigma}_\theta^2(\mathbf{z}_t,t)\Bigr).
\end{aligned}
\end{equation}

Applying Bayes’ rule to the forward transitions yields the exact posterior mean
\begin{equation}
\boldsymbol{\mu}_t(\mathbf{z}_t,\mathbf{z}_0)
= \frac{\sqrt{\alpha_t}(1-\bar\alpha_{t-1})}{1-\bar\alpha_t}\,\mathbf{z}_t
  + \frac{\sqrt{\bar\alpha_{t-1}}\,\beta_t}{1-\bar\alpha_t}\,\mathbf{z}_0,
\end{equation}
whose coefficients we denote by \(\mathcal{U}\) and \(\mathcal{E}\).  \texttt{BiMol-Diff}’s training objective is then to match the network’s predicted \(\boldsymbol{\mu}_\theta,\boldsymbol{\sigma}_\theta\) to these posterior quantities via a simple noise‐prediction loss. 
We optimize the negative log‐likelihood by upper‐bounding it with the variational lower bound
\begin{equation}
\mathbb{E}\bigl[-\log p_\theta(x_0)\bigr]
\;\le\;
\mathcal{L}_{\mathrm{vlb}}
\;=\;
\sum_{t=0}^T \mathcal{L}_t.
\end{equation}

\subsubsection{Variational Lower Bound (VLB)}

Following Sohl‐Dickstein et al.\cite{pmlr-v37-sohl-dickstein15}, for conditional generation the VLB decomposes into:
\begin{equation}
\begin{aligned}
\mathcal{L}_{\mathrm{vlb}}
&=
\mathbb{E}_{q(\mathbf{z}_{1:T}\mid\mathbf{z}_0)} \Bigl[
    \mathcal{L}_T
    + \sum_{t=2}^T \mathcal{L}_t
    - \mathcal{L}_0
\Bigr],\\
\mathcal{L}_T
&:= \log\frac{q(\mathbf{z}_T\mid \mathbf{z}_0)}{p(\mathbf{z}_T)},\\
\mathcal{L}_t
&:= \log
\frac{
    q(\mathbf{z}_{t-1} \mid \mathbf{z}_t,\mathbf{z}_0)
}{
    p_\theta(\mathbf{z}_{t-1}\mid \mathbf{z}_t,\mathbf{c})
},\\
\mathcal{L}_0
&:= \log p_\theta(\mathbf{z}_0 \mid \mathbf{z}_1,\mathbf{c}).
\end{aligned}
\end{equation}

\noindent where each $\mathcal{L}_t$ is a KL divergence between Gaussians.  
The true posterior mean (via Bayes' rule) is:
\begin{equation}
    \boldsymbol{\mu}_t(\mathbf{z}_t,\mathbf{z}_0)
    =
    \underbrace{
        \frac{\sqrt{\alpha_t}(1-\bar{\alpha}_{t-1})}
             {1-\bar{\alpha}_t}
    }_{\mathcal{U}}
    \mathbf{z}_t
    +
    \underbrace{
        \frac{\sqrt{\bar{\alpha}_{t-1}}\beta_t}
             {1-\bar{\alpha}_t}
    }_{\mathcal{E}}
    \mathbf{z}_0,
\end{equation}
with covariance $\boldsymbol{\Sigma}_q = \tilde{\beta}_t \mathbf{I}$, $\tilde{\beta}_t =
    \frac{1-\bar{\alpha}_{t-1}}{1-\bar{\alpha}_t}\beta_t.$
In the standard simplification, the model's covariance is fixed to match the true posterior covariance $\boldsymbol{\Sigma}_\theta = \boldsymbol{\Sigma}_q$,  
the KL collapses to a weighted MSE:
\begin{equation}
    \mathcal{L}_t
    = \frac{1}{2}
        \bigl\|
            \boldsymbol{\mu}_t
            - \boldsymbol{\mu}_\theta
        \bigr\|_{\boldsymbol{\Sigma}_q^{-1}}^2
    \;\propto\;
    \mathbb{E}\!
    \left[
        \bigl\|
            \mathbf{z}_0
            - \mathcal{M}_\theta(\mathbf{z}_t,t,\mathbf{c})
        \bigr\|^2
    \right].
\end{equation}
Thus, for $2\le t\le T$, $ \mathcal{L}_t
    \rightarrow
    \|\mathbf{z}_0
    - \mathcal{M}_\theta(\mathbf{z}_t,t,\mathbf{c})\|^2.$
The final KL encourages $\mathbf{z}_T$ to match the unit Gaussian prior:
\begin{equation}
    \mathcal{L}_T
    = \mathrm{KL}\bigl(q(\mathbf{z}_T\mid\mathbf{z}_0)\,\|\,p(\mathbf{z}_T)\bigr)
    \;\propto\;
    \bigl\|\boldsymbol{\mu}(\mathbf{z}_T)\bigr\|^2,
\end{equation}
a constant w.r.t.\ $\theta$.
The discrete target $\mathbf{S}$ (sequence) is encoded into a continuous embedding $g_\Phi(\mathbf{S})$.  
The final term in VLB is $\mathcal{L}_0 = -\log p_\theta(\mathbf{z}_0\mid\mathbf{z}_1, \mathbf{c})$. We need to integrate the discrete data $\mathbf{S}$ into this continuous likelihood term.
We use the law of total probability to express the continuous likelihood $p_\theta(\mathbf{z}_0\mid\mathbf{z}_1, \mathbf{c})$ by marginalizing over all possible discrete tokens in the target sequence $\mathbf{S} = \{s_1, s_2 \cdots, s_N\}$: 
\begin{equation}
    p_\theta(\mathbf{z}_0 \mid \mathbf{z}_1, \mathbf{c}) = \sum_{\mathbf{S}} p_\theta(\mathbf{z}_0, \mathbf{S} \mid \mathbf{z}_1, \mathbf{c})
\end{equation}
We then apply the product rule to the joint probability:
\begin{equation}
    p_\theta(\mathbf{z}_0, \mathbf{S} \mid \mathbf{z}_1, \mathbf{c}) = p_\theta(\mathbf{z}_0 \mid \mathbf{S}, \mathbf{z}_1, \mathbf{c}) \cdot p_\theta(\mathbf{S} \mid \mathbf{z}_1, \mathbf{c})
\end{equation}
For training, we are interested in the specific ground-truth sequence $\mathbf{S}$. When we evaluate $\mathcal{L}_0$ during training, we consider only the term where $\mathbf{S}$ is the ground-truth sequence:
\begin{equation}
\begin{aligned}
\mathcal{L}_0
&\approx -\log p_\theta(\mathbf{z}_0, \mathbf{S} \mid \mathbf{z}_1, \mathbf{c}) \\
&= -\log \Bigl[
    p_\theta(\mathbf{z}_0 \mid \mathbf{S}, \mathbf{z}_1, \mathbf{c})
    \cdot
    p_\theta(\mathbf{S} \mid \mathbf{z}_1, \mathbf{c})
\Bigr].
\end{aligned}
\end{equation}

\noindent The core approximation simplifies the dependency graph by asserting that the discrete data $\mathbf{S}$ is generated only from the clean latent $\mathbf{z}_0$, and is independent of $\mathbf{z}_1$ and $\mathbf{c}$ given $\mathbf{z}_0$.
\begin{equation}
    \mathbf{S} \perp (\mathbf{z}_1, \mathbf{c}) \mid \mathbf{z}_0
\end{equation}
This allows us to replace the discrete conditional likelihood with the separate rounding network $\tilde{p}_\Phi(\mathbf{S} \mid \mathbf{z}_0)$:
$p_\theta(\mathbf{S} \mid \mathbf{z}_1, \mathbf{c}) \approx \tilde{p}_\Phi(\mathbf{S} \mid \mathbf{z}_0)$.
Substituting this back into the likelihood decomposition:
\begin{equation}
    p_\theta(\mathbf{z}_0, \mathbf{S} \mid \mathbf{z}_1, \mathbf{c}) \approx p_{\mathrm{cont}}(\mathbf{z}_0 \mid \mathbf{S}, \mathbf{z}_1, \mathbf{c}) \cdot \tilde{p}_\Phi(\mathbf{S} \mid \mathbf{z}_0)
\end{equation}
Taking the negative logarithm of the approximation gives the two desired terms:

$$\mathcal{L}_0 \approx -\log p_{\mathrm{cont}}(\mathbf{z}_0 \mid \mathbf{S}, \mathbf{z}_1, \mathbf{c}) - \log \tilde{p}_\Phi(\mathbf{S} \mid \mathbf{z}_0)$$

\noindent This split yields the two components used in the final training objective:
\begin{enumerate}
    \item Consistency Term ($\mathcal{L}_{\mathrm{Cons}}$): The first term is the negative log-likelihood of the continuous latent, which is minimized via the MSE loss on the means:
    $-\log p_{\mathrm{cont}}(\mathbf{z}_0 \mid \mathbf{S}, \mathbf{z}_1, \mathbf{c}) \rightarrow \mathcal{L}_{\mathrm{Consistency}} = \bigl\| g_\Phi(\mathbf{S}) - \mathcal{M}_\theta(\mathbf{z}_1,1,\mathbf{c}) \bigr\|^2$.
    \item Rounding Term ($\mathcal{L}_{\text{Round}}$): This second term is the dedicated loss for the discrete data likelihood:
    $\mathcal{L}_{\text{Round}} = -\log \tilde{p}_\Phi(\mathbf{S} \mid \mathbf{z}_0)$
\end{enumerate}

\subsubsection{Final End-to-End Objective}

Combining all components:
\begin{align}
\mathcal{L}_{\mathrm{vlb}}
\propto\;
&\sum_{t=2}^T
\underbrace{
    \bigl\|
        \mathbf{z}_0
        - \mathcal{M}_\theta(\mathbf{z}_t,t,\mathbf{c})
    \bigr\|^2
}_{\text{Denoising}}
\\
&+
\underbrace{
    \bigl\|
        g_\Phi(\mathbf{S})
        - \mathcal{M}_\theta(\mathbf{z}_1,1,\mathbf{c})
    \bigr\|^2
}_{\text{Consistency}}
\\
&-
\underbrace{
    \log \tilde{p}_\Phi(\mathbf{S}\mid\mathbf{z}_0)
}_{\text{Rounding}}.
\end{align}

Dropping constant terms, the simplified end-to-end training loss is:
\begin{equation}
\begin{aligned}
\mathcal{L}_{\text{e2e-simple}}(\mathbf{S})
&=
\mathbb{E}_{q}
\Bigl[
    \sum_{t=2}^{T}
    \underbrace{
        \bigl\|
            \mathcal{M}_\theta(\mathbf{z}_t, t, \tilde{\mathcal{G}})
            - \mathbf{z}_0
        \bigr\|^2
    }_{\text{Denoising}}
\\
&\qquad+
    \underbrace{
        \bigl\|
            g_\Phi(\mathbf{S})
            - \mathcal{M}_\theta(\mathbf{z}_1, 1, \tilde{\mathcal{G}})
        \bigr\|^2
    }_{\text{Consistency}}
\\
&\qquad-
    \underbrace{
        \log \tilde{p}_\Phi(\mathbf{S} \mid \mathbf{z}_0)
    }_{\text{Rounding}}
\Bigr].
\end{aligned}
\end{equation}

\subsection{Non-increasing Isotonic Projection}\label{projection}
After constructing the per-token cumulative schedule $\tilde{\alpha}^i_t$, we project it onto the set of non-increasing sequences $\{\bar{\alpha}^i_t\}_{t=1}^T$ such that $\bar{\alpha}^i_1 \ge \bar{\alpha}^i_2 \ge \dots \ge \bar{\alpha}^i_T$. Concretely, this is a 1D isotonic regression problem with squared loss, which we solve using the standard Pool-Adjacent-Violators Algorithm (PAVA). This algorithm finds the closest monotone non-increasing sequence (in the least-squares sense) to the input. Intuitively, it smooths out spurious "bumps" in the loss profile while guaranteeing that the cumulative signal strength strictly decays over time, fulfilling the monotonicity requirement of the diffusion process.

\end{document}